\documentclass[10pt,conference]{IEEEtran}
\IEEEoverridecommandlockouts

\usepackage{cite}
\usepackage{amsmath,amssymb,amsfonts}
\usepackage{algorithmic}
\usepackage{graphicx}
\usepackage{hyperref}
\usepackage{textcomp}
\usepackage{tabularx}
\usepackage{multirow}
\usepackage{booktabs}
\usepackage[most]{tcolorbox}
\usepackage{xcolor}
\usepackage{adjustbox}
\def\BibTeX{{\rm B\kern-.05em{\sc i\kern-.025em b}\kern-.08em
    T\kern-.1667em\lower.7ex\hbox{E}\kern-.125emX}}
\begin{document}

\title{SKIP: a Self-knowledge-guided Step-wise Preference Learning Framework for Concise Reasoning\\
}
\author{
\IEEEauthorblockN{
Qinhong Lin\textsuperscript{a}, Yuhao Zhang\textsuperscript{a}, Yinglun Feng\textsuperscript{a}, Zhongliang Yang\textsuperscript{b,a,*}, Linna Zhou\textsuperscript{b,a,*}
}
\IEEEauthorblockA{
\textsuperscript{a}School of Cyberspace Security, Beijing University of Posts and Telecommunications, Beijing, China \\
\textsuperscript{b}QuanCheng Laboratory, Jinan, China
}
\IEEEauthorblockA{
\{greenred99, yangzl, zhoulinna\}@bupt.edu.cn
}
}

\maketitle

\begin{abstract}
While Chain-of-Thought (CoT) reasoning has been proven to be effective, it often leads to overthinking, resulting in computational overhead, inference latency, and even degraded performance in large language models (LLMs). Existing concise reasoning frameworks significantly compromise accuracy while compressing the length of output. In this paper, we propose SKIP, a self-knowledge-guided step-wise preference learning framework. Starting with lightweight fine-tuning to adjust the model’s output style, SKIP introduces a carefully designed knowledge probing mechanism to guide model to output an answer at each reasoning step. Based on the correctness of intermediate steps, we construct preference data that guide the model toward more efficient and correct reasoning by leveraging DPO. Experimental results demonstrate that our method effectively improves reasoning compression while mitigating performance degradation after fine-tuning. Besides, SKIP shows strong generalization ability on out-of-distribution datasets. We further conducted ablation studies on the component parameters of our framework.
\end{abstract}

\begin{IEEEkeywords}
consise reasoning, LLM, DPO, CoT.
\end{IEEEkeywords}

\section{Introduction}
Chain-of-Thought (CoT) reasoning has substantially enhanced the reasoning capabilities of large language models (LLMs), improving their performance on complex tasks such as mathematical reasoning \cite{c1}. This advancement holds great promise for enabling LLMs and agents to participate in human decision-making as powerful assistants \cite{c2}. Despite these advantages, deeper investigations into the mechanism of CoT reasoning have revealed that LLMs tend to engage in overthinking \cite{c3}. This tendency not only incurs additional computational overhead and inference latency but also undermines performance when the model
is confident to answer question directly. 
\begin{figure}[htbp]
    \centering
    \includegraphics[width=0.7\linewidth]{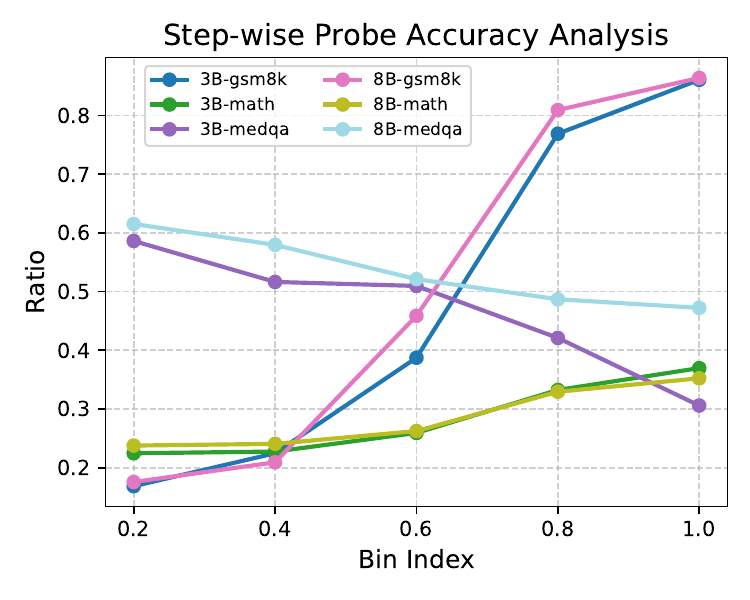} 
    \caption{Step-wise probe accuracy analysis across reasoning progress. Bin Index represents various stages of the inference process, relative to the complete inference. In the GSM8K and Math datasets, the reasoning exhibits positive returns, while in MedQA, it shows negative returns.}
    \label{fig:ratio}
\end{figure}
Reference \cite{c28}'s comparative experiments on multiple tasks show that the gains brought by COT are related to the problem type.
Empirical studies further show that LLMs exhibit a preference for lengthy answers\cite{c4}, suggesting that they lack inherent constraints to avoid redundancy or encourage concise reasoning. 
Thus, various studies have been proposed that focus on compressing reasoning length without compromising model performance\cite{c5}.
Existing research can be broadly categorized into the following two directions. One is inference-time intervention which tries to add control mechanisms like budget constraints or internal state probes at the model inference process.
Another one is model capability elicitation. This line of work aims to encourage concise reasoning through data construction and training objectives.
The training approaches, through sophisticated data construction and training, endows the model with the ability to simplify inference, making it more flexible without causing additional inference latency, and this capability can be transferred between different datasets.
However, previous work mainly focuses on training models using data guided by context engineering or sampled from more powerful models. This drastic style shift may lead to length compression while severely compromising performance. 
In this study, we also focus on the latter direction to compress reasoning trace, aiming to
tackle three core questions:
\textbf{Q1:} How to efficiently construct data to elicit the model’s ability to concisely compress reasoning?
\textbf{Q2:} How can we prevent the drop of reasoning accuracy during compression?
\textbf{Q3:} Can this ability generalize to out-of-distribution (OOD) datasets?

\begin{figure*}[htbp]
    \centering
    \includegraphics[width=\linewidth]{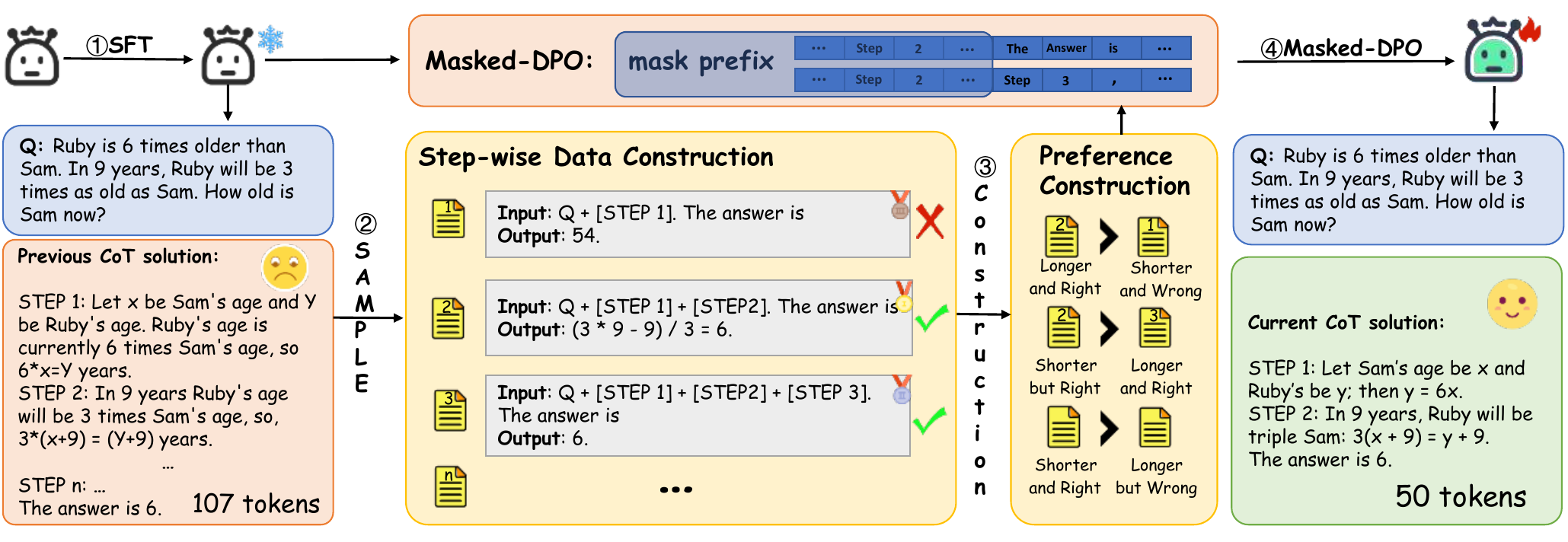} 
    \caption{\textbf{S}elf-\textbf{K}nowledge-gu\textbf{I}ded step-wise \textbf{P}reference learning (SKIP), a three stage reasoning compression framework. We use a simple yet efficient probe to detect whether the model can answer the question. We then train models with preference data to teach them to learn when to exit at specific points.}
    \label{fig:framework}
\end{figure*}

To tackle these questions, we propose leveraging the model’s own knowledge to construct efficent datasets for learning which facilitates reasoning compression. 
Our motivation is driven by a key observation: when models are forced to decode an answer, such as when they encounter the signal 'The answer is', they may be able to generate a correct response in certain situations."
As illustrated in Fig. \ref{fig:framework}, when 'The answer is' is inserted at the first step of reasoning and the model is forced to generate an answer, it fails to provide the correct one because the necessary reasoning elements are not yet available.
When “The answer is” is inserted at the second step, differently, the model is able to arrive at the correct answer after performing only simple calculations. Without such forced intervention, the model tends to continue producing redundant reasoning steps, thereby increasing the token budget. 
This phenomenon that some current reasoning paths indeed contain redundant steps, indicates that the model lacks the ability to recognize whether sufficient knowledge has been acquired during reasoning, and thus fails to terminate the reasoning process at the appropriate stage.
As shown in Fig.\ref{fig:ratio}, we analyzed the step-wise probe accuracy on GSM8K, MATH, and MedQA with Llama3.1-8B-instruct and Llama3.2-3B-instruct. On reasoning benchmarks (GSM8K, MATH), the likelihood of a correct response correlates positively with CoT length, though with diminishing returns in the final stages. In stark contrast, the experimental results on MedQA exhibit a negative trend, with accuracy deteriorating as reasoning proceeds, as models lack sufficient domain knowledge. These findings underscore the necessity of a self-knowledge probing mechanism capable of dynamically identifying the sufficiency of reasoning to determine the optimal early-exit boundary.
In this work, we propose a \textbf{S}elf-\textbf{K}nowledge-gu\textbf{I}ded step-wise \textbf{P}reference learning (SKIP) framework. 
At a high level, it's a SFT-DPO two-stage training framework.
We first design a simple yet efficient probe to identify model inference state after cold start and construct reasoning compression data for LLMs without relying on external knowledge. 
We then carefully mix sampled contrastive data to guide the model through preference learning, enabling the compression of reasoning paths without degrading reasoning performance.
Experiments on both In-Distribution (ID) and Out-Of-Distribution (OOD) datasets show that the proposed method not only shortens reasoning length effectively but also improves the reasoning capability of LLMs.
Our main contributions are as follows:
\begin{itemize}
    \item We propose a model-free reasoning boundary probing mechanism that efficiently determine whether the model already has the ability to output the correct answer.
    \item We propose a data construction scheme that balances knowledge preference and length preference, which can be used to train models to improve inference length compression capabilities while mitigating performance degradation under direct preference optimization training.
    \item Extensive experiments across different tasks and models prove the robutness of SKIP. Besides, we conduct ablation studies on the training data to disentangle and analyzed its specific impact on model performance.
\end{itemize}
Our data/code is available at \url{https://github.com/linqinhong/SKIP}.

\section{Related Work}
\noindent \textbf{Chain-of-Thought Reasoning.}
Chain-of-Thought (CoT)\cite{c24} reasoning has proven to be a transformative paradigm for achieving human-like cognition and solving complex tasks, as exemplified by the success of reasoning-centric models such as OpenAI o1\cite{c25}, Gemini-3.0-Pro\cite{c26}, and DeepSeek-R1\cite{c27}. During model training, CoT is integrated with techniques like Best-of-N (BoN)\cite{c29} and Monte Carlo Tree Search (MCTS)\cite{c30}, along with the variant, Tree-of-Thought (ToT), to systematically explore the distribution space and sample high-quality synthetic data. Furthermore, the paradigm shift in Reinforcement Learning (RL) from outcome-based rewards\cite{c29} to process-based rewards\cite{c31} highlights how step-by-step reasoning facilitates deeper cognition, pushing model applications into more specialized and sophisticated domains.

\noindent \textbf{Concise Reasoning.}
Concise Reasoning aims to mitigate the prohibitive computational overhead and the "overthinking" problem.
One primary approach involves inference-time interventions to manage computational budgets. For instance, soft control methods, such as TALE\cite{c6}, inject budget constraints into the prompt to guide the model toward shorter reasoning paths. In contrast, hard control strategies, such as \cite{c7,c33}, employ a prober to predict whether explicit reasoning is necessary for a given task based on the internal states.
Another line of work focuses on inducing more concise reasoning styles within the model itself. For instance, FS-BoN \cite{c8} leverages few-shot examples to demonstrate brevity, while C3OT \cite{c9} utilizes proprietary APIs to distill and compress reasoning steps. These methods typically involve fine-tuning the model to shift its inherent reasoning trajectory toward conciseness. Other studies such as \cite{c10,c11,c12} design length-constrained reward functions within Reinforcement Learning (RL) frameworks to penalize excessive verbosity.

Building on these insights, our work leverages internal state probing to facilitate the construction of high-quality training data. We introduce an explicit mechanism to identify knowledge-sufficient states, enabling the model to strategically bypass redundant reasoning trajectories when its internal information is already adequate.

\section{Methodology}
\label{sec:Methodology}
As illustrated in Fig. \ref{fig:framework}, the SKIP framework consists of three stages. First, we leverage a small set of answers with concise reasoning to supervised fine-tune (SFT) the model to learn a more succinct response style. Next, we employ a step-wise knowledge probing mechanism to determine whether the model has already acquired sufficient knowledge to produce the correct answer, and leverage this assessment as guidance for preference-based data construction. Finally, we apply masked-DPO training to inject this capability into the model.

\subsection{SFT Cold Start}\label{SFT_cold_start}
Unaligned models often produce answers with inconsistent formats. To address this issue, we first perform SFT cold-start training with two objectives: (i) encouraging the model to output a final summary answer and terminate unnecessary reasoning whenever “The answer is” appears at the end of a reasoning step, and (ii) guiding the model to adopt a more concise response style, thereby providing a better foundation for subsequent DPO training.
Specifically, we begin with $N=8$ question–answer pairs that end with a “The answer is” format as few-shot conditions, and employ in-context learning (ICL)\cite{c23} to elicit the model to generate additional pairs in the same style. We filtered data according to correctness and then fine-tune the model via SFT to internalize this reasoning style.
In practice, we found that this approach effectively achieves the intended objectives, but also reveals a trade-off between efficiency and reasoning capability: while the reasoning path is compressed, the model’s reasoning performance may experience slight degradation.


\subsection{Self-Knowledge-Guided Preference Data}
\label{skip}
\noindent\textbf{Inference Boundary Probing.}
After supervised fine-tuning, the model becomes more inclined to output the final answer following a “The answer is” signal. 
The key to effectively reducing the model's reasoning length while minimizing performance degradation lies in the model's ability to recognize when the generated reasoning path is complete. To achieve this, we employ a simple and effective probe to guide the model within its sampling space. Specifically, during the reasoning process, we use 
a probe to guide the model to decode and output an answer based on its existing reasoning.
So, we sample candidates $Y=\{y_0,y_1,...y_i\}$ for the question Q. Each candidate can be formulate as:
\begin{equation}
    y_i=M(Q,t_0,t_1,...,t_{i-1},\boxed{probe})
\end{equation}
where i denotes the timestep and t denotes the token.
Greedy decoding serves as a reliable proxy for the model's explicit reasoning ability. We approximate a greedy-decoded answer as a representation of the model's internal state. We then assess the correctness of $y_i$ to determine whether the model has acquired sufficient reasoning elements (without confusion, all answers guided by probes are generated through greedy decoding). Therefore, producing the correct answer in this setting indicates that the model has acquired a stable answering capability.
Technically, inserting probes at any position is permissible. However, considering that a sentence is the smallest effective unit of the model's reasoning chain, we only insert probes after each sentence. Therefore, the input format to the model can be represented as: "$Q \backslash n [Step_1] \backslash n [Step_2] \backslash n ... \backslash n [Step_i] \backslash n \boxed{probe}$". 
The \boxed{probe} can be any phrase that prompts the model to output an answer. We use 'The answer is' to align with the cold start template, ensuring output quality.

Though models almost always produce a correct answer when the preceding reasoning steps are already logically sufficient, it's important to note that the ability to decode the correct answer at an early step does not guarantee that subsequent reasoning steps will remain correct.This may be because the model, in some cases, has already completed the reasoning implicitly in its internal representations, and may introduce instability or cause logical errors to accumulate when generate additional explicit steps\cite{c13}. Please refer to App. \ref{sec:case_study}.

\noindent \textbf{Step-wise Preference.} Our objective of this stage is twofold: (i) to preserve the reasoning capability of the model as much as possible, and (ii) to encourage the model to terminate reasoning once the existing steps are sufficient to produce the correct answer.
As illustrated in Fig. \ref{fig:framework}, we categorize the sampled answers into four types based on their correctness and length: $longer\ and\ right$, $shorter\ and\ wrong$, $longer\ but\ wrong$ and $short\ and\ right$. 
In this work, we focus on collecting preference data from two scenarios: knowledge preference data and length preference data.
When the model produces a correct answer at $step_i$ but fails to do so at one of $\{step_{t}|t<i\}$, this indicates that the model has strengthened its reasoning in the additional steps, which represents effective reasoning. We want the model to learn the correct reasoning logic from this. Therefore, we construct knowledge preference data with the relation $longer\ and\ right \succ shorter\ and\ wrong$.
When the model successfully produces a correct answer at $step_i$ and also does so at one of $\{step_{t}|t>i\}$, this indicates that both reasoning paths are valid. In this case, the extra reasoning in the longer answer should be considered redundant, and we want the model to learn a more concise reasoning pattern from this comparison. Thus, we construct length preference data with the relation $longer\ and\ right < shorter\ but\ right$.
In our experiments, we found that the length preference data has a significant impact on the model, leading to a rapid compression of length but also causing a sharp decline in performance. Therefore, we consider a correct answer at $step_i$ only when it is accompanied by an incorrect answer at $step_{t<i}$ and a correct answer at $step_{t>i}$. We analyze this situation in the App. \ref{app:Length Preference Data}.
Additionally, the reverse case of knowledge preference data, where the longer answer is wrong and the shorter one is correct, suggests that the additional reasoning steps caused an error in an otherwise correct answer. We want the model to terminate the reasoning process in a timely manner. Therefore, we treat $longer\ but\ wrong \succ shorter\ but\ wrong$ as a form of length preference data. It is worth noting that we found the sample size for this case to be very small.
It is important to note that we discard the $long\ but\ wrong\ VS\ short\ and\ wrong$ data, as it does not provide effective signals for training.


\noindent\textbf{Masked-DPO}. Given x representing the question, $y_w$ representing the preferred data and $y_l$ representing the rejected data,
we fine-tuned models using Direct Preference Optimization (DPO) \cite{c14} with the following loss:
\begin{equation}
    \mathcal{L}(\pi_\theta,\pi_{ref})=-\mathbb{E}_{D}[\beta\log\sigma(\log(\frac{\pi_\theta(y_w|x)}{\pi_{ref}(y_w|x)}/\frac{\pi_\theta(y_l|x)}{\pi_{ref}(y_l|x)}))]
\end{equation}
where there is $(x,y_w,y_l)\sim D$.
During training, we apply a masking strategy to the shared parts of the reasoning steps and question content between the chosen and rejected samples. By training only on the non-overlapping segments, we improve the stability of the optimization process.
To improve training efficiency, we fine-tune only the LoRA \cite{c15} parameters.
After training, we merged LoRA with the models for inference, which introduce no inference latency.

\section{Experiments}
\label{sec:experiment}
\noindent\textbf{Models and Datasets.}
We conducted experiments with Llama-3.1-8B-Instruct \cite{c16}, Llama-3.2-3B-Instruct\cite{c17} and Qwen-3-14B\cite{c35} models on two mathematical reasoning benchmarks that require multi-step reasoning to reach the final solution: GSM8K\cite{c18} and MATH\cite{c19} and one domain dataset MedQA\cite{c34}. To further evaluate the generalization ability of our method, we also performed out-of-distribution (OOD) evaluations on StrategyQA\cite{c20} and the Date Understanding task from BIG-Bench Hard\cite{c21}.
In our experiments, we used the full GSM8K dataset (7,473 training and 1,319 test examples). For MATH, we trained on all 7,500 examples and tested on the first 500. We evaluated StrategyQA on a 500-example subset and BBH’s Date Understanding task on 250 test cases.

\noindent\textbf{Baselines.}
We adopted one inference-time intervention method, Estimated Budget method\cite{c6}. For model capbility elicitation methods, we considered the Best-of-N (BON) sampling methods in \cite{c8} and the C3OT method in \cite{c9}. BON generated training fine-tuning data through few-shot prompting, while C3OT leveraged a stronger closed-source model to compress sampled reasoning trajectories for training. We adopted a prompt-guided CoT approach \cite{c22} without any additional control as our default baseline.
To better demonstrate the effectiveness of our self-knowledge-guided preference data, we adopted the BON training pipeline as the cold-start setting with $N=1$ and $N=8$.
While BON utilized the entire training set during SFT, we only used half of the data for SFT and constructed preference pairs from the remaining half.
For each setting, we ran the experiments five times and report the mean results.

\noindent\textbf{Evaluation Metrics.}
Our evaluation metrics included final answer accuracy and average reasoning length. Our goal was to achieve greater reasoning compression with minimal loss in accuracy. 
In our experiments, we judge answer via regular expression extraction and Exact Match evaluation.
We report the consistency of the evaluation results between this approach and LLM-as-a-Judge\cite{c32} in the App. \ref{app:LLM-as-a-judge}.
We also compared the compression efficiency defined as:
\begin{equation}
    Eff =\frac{Accuracy}{Average\ Length},
\end{equation}
which calculate the accuracy brought by each token.

\begin{table*}[htbp]
\centering
\label{tab:main-result}
\caption{Results of different reasoning compression methods. The bold text represents the best result for the corresponding metric in the block, and the text with a \textsuperscript{\textdagger} represents the second best result for the corresponding metric in the block.}
\begin{tabular}{c|c|ccc|ccc|ccc}
\toprule
\multirow{2}{*}{model}     &    \multirow{2}{*}{method}    & \multicolumn{3}{c|}{GSM8K} & \multicolumn{3}{c|}{Math} & \multicolumn{3}{c}{MedQA}                                           \\
 &  & Acc     & Length  & Eff   & Acc    & Length  & Eff   & Acc                          & Length                        & Eff  \\
\midrule
    & Default                  & 76.70   & 217.70  & 0.35  & \textbf{47.20}  & 503.20  & 0.09  & 56.83 & 482.36 & 0.12 \\
    & Budget                   & 72.30   & 141.50  & 0.51  & 44.60  & 438.20  & 0.10  & 40.35                        & 339.80                        & 0.12 \\
    & C3OT                     & 68.80   & 115.30\textsuperscript{\textdagger}  & 0.60  & 39.60  & \textbf{320.50}  & 0.12  & 54.79                        & 101.80                        & 0.54 \\
    & BON-1                    & 77.30   & 149.30  & 0.52  & 43.60  & 392.00  & 0.11  & 56.83                        & 99.14                         & 0.57 \\
    & SKIP-1                   & 77.70   & 120.10  & 0.65\textsuperscript{\textdagger}  & 45.30\textsuperscript{\textdagger}  & 376.30  & 0.12  & 58.08                        & 97.47                         & 0.60 \\
    & BON-8                    & 78.90\textsuperscript{\textdagger}   & 130.30  & 0.61  & 44.20  & 333.00  & 0.13\textsuperscript{\textdagger}  & \textbf{58.50}                        & 83.00\textsuperscript{\textdagger}                         & 0.70\textsuperscript{\textdagger} \\
\multirow{-7}{*}{Llama-3.2-3B}    & SKIP-8                   &\textbf{ 79.20}   & \textbf{114.00}  & \textbf{0.69}  & 43.80  & 321.70\textsuperscript{\textdagger}  &\textbf{ 0.14}  & 58.18\textsuperscript{\textdagger}                       &  \textbf{74.19}                         & \textbf{0.80} \\
\midrule
                        & Default                  & \textbf{85.50}   & 240.40  & 0.36  & 46.60  & 479.50  & 0.10  & 52.70                        & 278.00                        & 0.19 \\
                        & Budget                   & 80.40   & 149.50  & 0.54  & 46.20  & 440.20  & 0.10  & 56.51                        & 433.92                        & 0.13 \\
                        & C3OT                     & 77.20   & \textbf{109.30}  & 0.71\textsuperscript{\textdagger}  & 37.40  & \textbf{302.70}  & 0.12\textsuperscript{\textdagger}  & 66.30                        & 110.80                        & 0.60 \\
                        & BON-1                    & 81.60   & 154.00  & 0.53  & 46.30  & 418.50  & 0.11  & 67.50                        & 87.50                         & 0.77 \\
                        & SKIP-1                   & 82.90   & 115.20\textsuperscript{\textdagger}  & \textbf{0.72}  & \textbf{47.80}  & 398.20  & 0.12\textsuperscript{\textdagger}  & 67.66\textsuperscript{\textdagger}                        & 87.51                         & 0.77 \\
                        & BON-8                    & 83.40   & 127.10  & 0.66  & 45.10  & 374.30  & 0.12\textsuperscript{\textdagger}  & 67.00                        & 76.00\textsuperscript{\textdagger}                         & 0.88\textsuperscript{\textdagger} \\
\multirow{-7}{*}{Llama-3.1-8B}    & SKIP-8                   & 84.35\textsuperscript{\textdagger}   & 119.00  & 0.71\textsuperscript{\textdagger}  & 46.70\textsuperscript{\textdagger}  & 367.90\textsuperscript{\textdagger}  & \textbf{0.13}  & \textbf{68.90}                        & \textbf{73.10}                         & \textbf{0.94} \\
\midrule
                        & Default                  & 88.93   & 473.57  & 0.19  & 70.60  & 983.48  & 0.07  & 59.65                        & 958.50                        & 0.06 \\
                        & Budget                   & 82.18   & 359.94  & 0.23  & 41.10  & 855.60  & 0.05  & 47.10                        & 685.54                        & 0.07 \\
                        & C3OT                     & 86.84   & \textbf{132.84}  & 0.65\textsuperscript{\textdagger}  & 65.40  & 700.47  & 0.09  & 71.74                        & 439.67                        & 0.17 \\
                        & BON-1                    & 93.86\textsuperscript{\textdagger}   & 187.48  & 0.50  & 76.20  & 513.25  & 0.15  & 72.53  \textsuperscript{\textdagger}                      & 255.00                        & 0.28 \\
                        & SKIP-1                   & \textbf{95.45}   & 152.47  & 0.63  & \textbf{84.40}  & 359.00\textsuperscript{\textdagger}  & 0.24\textsuperscript{\textdagger}  & \textbf{72.90}                        & \textbf{164.00}                        & 0.31 \\
                        & BON-8                    & 92.65   & 161.02  & 0.58  & 75.40  & 487.45  & 0.15  & 71.74                        & 224.44                        & 0.32\textsuperscript{\textdagger} \\
\multirow{-7}{*}{Qwen3-14B}   & SKIP-8                   & 93.27   & 138.28\textsuperscript{\textdagger}  & \textbf{0.67}  & 81.20\textsuperscript{\textdagger}  & \textbf{319.00}  & \textbf{0.25}  & 71.80                        & 190.00\textsuperscript{\textdagger}                        & \textbf{0.38} \\
\bottomrule
\end{tabular}
\end{table*}

\begin{table}[t]
  \centering
  \caption{OOD experiment results on Llama-3.1-8B. The bold scores denote the best performance.}
  \label{tab:ood-experiment}
  \begin{tabular}{cl|llc|llc}
    \toprule
    \multirow{2}{*}{Train\ data}  &  \multirow{2}{*}{Method} & \multicolumn{3}{c|}{BBH(DU)} & \multicolumn{3}{c}{StrategyQA} \\ 
      &   & Acc  & Len & Eff & Acc  & Len & Eff\\
    \midrule
    & Default & 76.8  & 257.5 & 0.298 & 69 & 303.8  & 0.227  \\ 
    \midrule
    \multirow{4}{*}{GSM8K}
           & BON1  & 66.9 & 125.2 & 0.534  & 51.3 & 156.5 & 0.328 \\ 
           & SKIP-1  & \textbf{72.2} & 141.7 & 0.509  & 63.1 & \textbf{124.8} & \textbf{0.505} \\ 
           & BON8  & 62.4 & 88.3 & 0.706 & 56.3 & 142.3 & 0.395 \\ 
           & SKIP-8  & 67.0 & \textbf{85.1} & \textbf{0.787}  & \textbf{64.1} & 127.9 & 0.501  \\ 
    \midrule
    \multirow{4}{*}{MATH}
           & BON1  & 67.4 & 147.6  & 0.456 & 57.5 & 241.3 & 0.23      \\ 
           & SKIP-1 & 68.0 & 143.5 & 0.473  & 59.8 & 238.8 & 0.250  \\ 
           & BON8 & 69.0 & 93.6 & 0.737 & 65.1 & \textbf{157.7} & 0.412  \\ 
           & SKIP-8 & \textbf{70.8} & \textbf{91.8} & \textbf{0.771}  & \textbf{65.3} & 158.0 & \textbf{0.413}  \\ 
    \bottomrule
  \end{tabular}
\end{table}

\subsection{ID results and OOD results}
\label{ssec:subhead}
In Tab. \ref{tab:main-result}, We report the results on the in-distribution (ID) datasets. We summarize the key findings as follows: 

(1) Our SKIP-8 achieves the highest token efficiency in nearly all experiments, except for one setup where it ranks second. All finetuning methods effectively compress the reasoning length of LLMs compare to the Estimated Budge method. Moreover, in all experiments, both SKIP-1 and SKIP-8 show significant improvements compared to BON-1 and BON-8 with similar training data sizes. Our results show that SKIP framework directly address \textbf{Q1:} How to efficiently construct data to elicit the model’s ability to concisely compress reasoning?
We show a case study in Appendix~\ref{fig:case study}.

(2) Two baselines other than BON are also able to reduce reasoning length. However, this comes at the cost of a significant drop in accuracy. By incorporating knowledge preference data, SKIP not only compresses reasoning more effectively but also improves accuracy, providing a positive answer to \textbf{Q2:} How can we prevent the drop of reasoning accuracy during compression? It demonstrates that combining knowledge preference data with length preference data is a simple, black-box, yet efficient method.

(3) C3OT could achieves a higher compression ratio than BON methods in some settings, particularly on the GSM8K and Math dataset with Llama-3.1-8B. We refer it to the reason that C3OT leverages API-based sampling with powerful closed-source models to identify redundant reasoning steps during training, thereby achieving strong compression. 
By contrast, BON relies on in-context learning to guide concise reasoning, which is, to some extent, constrained by the capacity of the base model. This limitation becomes more apparent on challenging datasets like MATH, where C3OT clearly outperforms in terms of compression. However, there is a sharp decline in accuracy with C3OT, indicating that training data from other models may not align well with the model's historical knowledge and capabilities. This suggests that transferring expression styles incurs greater costs. Unlike BON, SKIP utilizes data from distribution sampling, and does not show any significant deterioration in correctness.

In Tab. \ref{tab:ood-experiment}, we presented results on out-of-distribution (OOD) datasets. We exclude C3OT and Estimated Budget from this comparison due to their substantial performance degradation on ID datasets, which diminishes the value of extending them to OOD scenarios. The key findings are: 

(1) The concise reasoning ability acquired through SFT generalizes effectively to OOD datasets. However, this also comes with a notable drop in accuracy compared to the Default model. 

(2) SKIP consistently outperforms BON across all three evaluation metrics. This demonstrates that our constructed preference data not only encourages concise reasoning but also enhances the model’s reasoning capability, even in OOD scenarios. These two provide an answer to our research question \textbf{Q3:} Can this ability generalize to out-of-distribution (OOD) datasets?



\subsection{Ablation Study}
In this section, we analyze the impact of the proposed preference data by examining the effects of data mixing ratios, training data size, and masked-DPO on model performance.

\begin{figure}[t]          
  \centering
  \includegraphics[width=\linewidth, trim=0 0 0 0, clip]{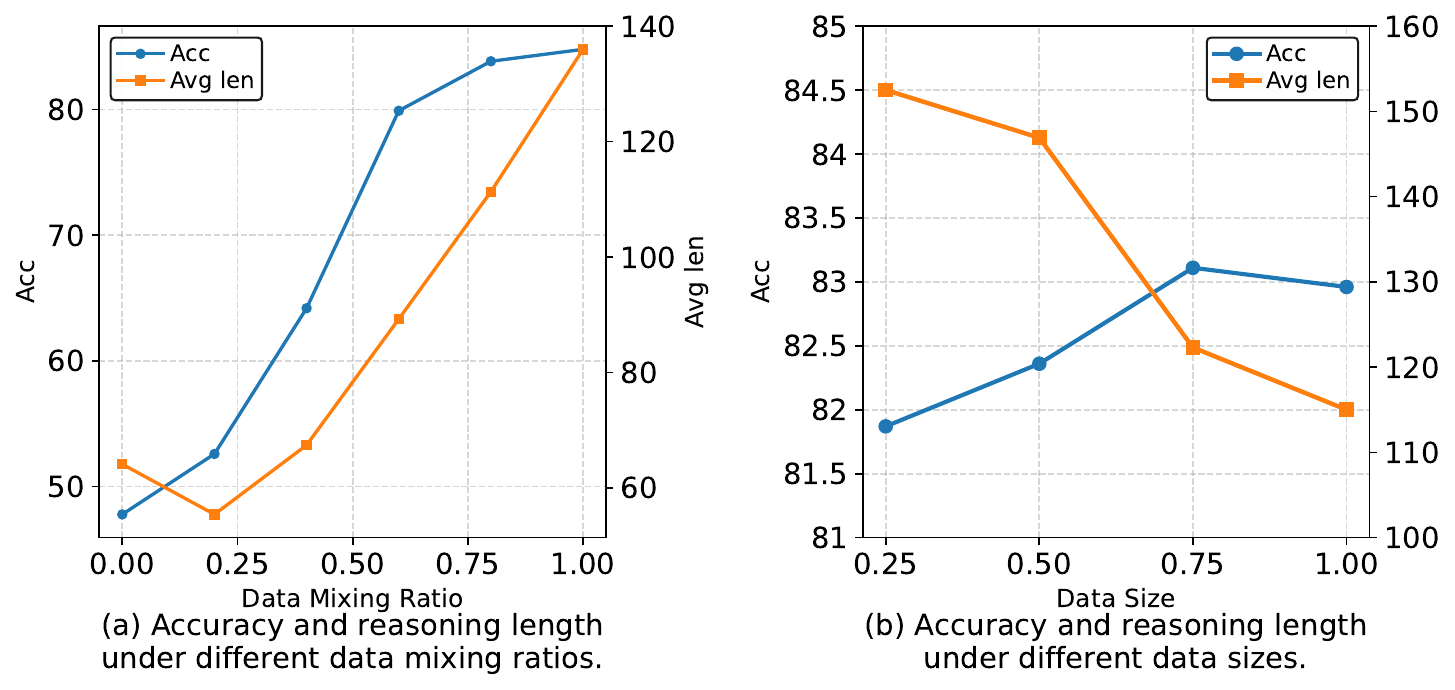}  
  \caption{a.We compared different proportions of knowledge preference data and length preference data to analyze their impact on model performance. b.We analyze the impact of training data size on model performance.}
  \label{fig:T2}
\end{figure}

\noindent\textbf{Data mixing ratio}
In Sec. \ref{skip}, we categorize the constructed data into knowledge preference data and length preference data based on forced decoding and the correctness of intermediate reasoning steps. In this section, we explicitly examine their respective effects on model accuracy and reasoning length after training. As shown in Fig. \ref{fig:T2}(a), increasing the proportion of knowledge preference data in the mixture leads to longer reasoning chains and higher accuracy. Knowledge preference data is derived from detecting incorrect outputs in forced decoding at earlier steps, encouraging the model to learn that sufficient knowledge must be accumulated before producing an answer. In contrast, length preference data plays the opposite role: it guides the model to avoid overthinking once enough knowledge has been acquired. Together, these two types of data form an adversarial learning process that balances reasoning sufficiency with conciseness.


 \begin{table}[t]
  \centering
  \caption{The difference between DPO and masked-DPO on Llama-3.1-8B. Masked-DPO could achieve better results.}
  \label{tab:DPO}
  \begin{tabular}{cl|ll|ll}
    \toprule
    \multirow{2}{*}{Method}  &  \multirow{2}{*}{DPO-type} & \multicolumn{2}{c|}{GSM8K} & \multicolumn{2}{c}{MATH} \\ 
      &   & Acc  & Len & Acc  & Len\\
    \midrule
    \multirow{2}{*}{SKIP-1}
           & DPO  & 82.18 & 145 & 47.1  & 412 \\ 
           & Masked-DPO  & \textbf{82.31} & \textbf{142} & \textbf{47.8}  & \textbf{401} \\ 
    \midrule
    \multirow{2}{*}{SKIP-8}
           & DPO  & 83.96 & 119  & 45.8 & 367.7   \\ 
           & Masked-DPO & \textbf{84.35} & 119 & \textbf{46.68}  & \textbf{366.9} \\ 
    \bottomrule
  \end{tabular}
\end{table}

\noindent\textbf{Training data size.}
In Fig. \ref{fig:T2}(b), we analyze the impact of training data size on the final performance. Specifically, we use 25\%, 50\%, 75\%, and 100\% of the half portion for preference data, as described in the experiment setup. As the amount of data increases, the model’s accuracy gradually improves while the length decreases steadily with more data. 
This aligns with the trend in model training and also reflects that our data effectively balances reasoning length and accuracy.

\noindent\textbf{Training Objective.}
In the constructed step-wise preference data, the chosen and rejected samples share a common prefix in addition to the question itself. To prevent this shared prefix from interfering with training, we apply a masking strategy during DPO training to exclude it. In Tab. \ref{tab:DPO}, we compare the two training settings. Experiments on LLaMA-3.1-8B show that masked-DPO yields shorter lengths and higher accuracy.

\section{Conclusion}
\label{sec:conclusion}
In this work, we presented SKIP, a self-knowledge-guided step-wise preference learning framework that leverages intermediate reasoning signals to construct knowledge and length preference data.By combining SFT cold start with step-wise probing, SKIP efficiently detects the model's reasoning critical points and constructs preference data for DPO training, enabling the model to learn more efficient concise reasoning abilities to generate shorter reasoning chains without sacrificing, and even improving, answer accuracy. Our experiments further demonstrate that this capability generalizes effectively to out-of-distribution datasets.

\section{Limitations}
Although we have demonstrated that SKIP exhibits stronger capabilities in concise reasoning, we acknowledge that there are areas in this paper that need further enhancement, which we leave to future work.

\noindent \textbf{More Robust Efficiency Metrics}: In this paper, we measure the model's compression efficiency by calculating the accuracy contribution of each individual token. However, we believe that the difficulty of improving model accuracy is influenced by diminishing returns—the closer the model approaches its performance threshold, the harder it becomes to achieve further improvements. The same applies to length compression. Therefore, finding a balance between both factors for a more robust analysis of the model's concise reasoning capability is an important direction for future research.

\noindent \textbf{Broader Evaluation}: 
Although the evaluation in this work covers a broad range, including various types and topics of test questions, it is necessary to apply and evaluate the method on a wider range of data.

\section*{acknowledgement}
This work was supported in part by Research Project of Quan Cheng Laboratory, China (Grant No.QCL20250203)and in part by the National Natural Science Foundation of China under Grant 62172053 and Grant 62302059.

\vspace{12pt}
\appendix

\subsection{LLM-as-a-Judge VS. Exact Match}
\label{app:LLM-as-a-judge}
In evaluating short answers, LLM-as-a-Judge is increasingly becoming a more effective and accurate approach for assessing answer correctness. 
Therefore, in Tab. \ref{tab:appendix-LLM-as-a-judge-tab}, we present the consistency score of the Llama-3.2-3B and Llama-3.1-8B models across three datasets between using both Exact Match (EM) and LLM-as-a-judge as evaluation methods.
For each dataset, we randomly sample 300 samples for comparison. The high agreement rate observed between the two methods supports the validity and reliability of our Regex-based EM evaluation pipeline.

\begin{table}[htbp]
\centering
\caption{The match ratio between EM score and LLM-as-a-judge on the SKIP8 experiment setting. High consistency supports the accuracy of the EM results.}
\label{tab:appendix-LLM-as-a-judge-tab}
\begin{tabular}{c|c|c}
\toprule
Model & Dataset      & Consistency \\
\midrule
\multirow{3}{*}{3B}    & GSM8K        & 0.91        \\
      & Math         & 0.89        \\
      & StrategyQA   & 0.87       \\
\midrule
\multirow{3}{*}{8B}    & GSM8K        & 0.94        \\
      & Math         & 0.87        \\
      & StrategyQA   & 0.85        \\

      \bottomrule
\end{tabular}
\end{table}
Furthermore, we report the evaluation results when LLM-as-a-judge is employed for data construction and training. The SKIP-1 method on Llama-3.1-8B improves the Best-of-N baseline from 81.6 / 154.0 / 0.53 to 82.1 / 101.7 / 0.807 when trained with data constructed through EM and to 83.47 / 138 / 0.6048 when trained with data constructed through LLM-as-a-Judge. We analyze the lower accuracy scores observed under EM compared to LLM-as-a-judge and attribute this to the fact that EM is prone to generating more False Negatives (i.e., judging a correct answer as wrong due to formatting). Consequently, the LLM-as-a-judge approach successfully identifies more valid preference pairs satisfying "$Longer\ \&\ Right \succ Shorter\ \&\ Wrong$". This enrichment of the training data encourages the model to acquire correct answers through longer, more robust reasoning processes. 

\subsection{Length Preference Data}
\label{app:Length Preference Data}
As discussed in Sec. \ref{skip},
we define a valid data instance at $step_i$ specifically when it is accompanied by an incorrect answer at $step_{t<i}$ and a correct answer at $step_{t>i}$. Using Llama-3.1-8B on the GSM8K and MATH datasets, we investigate how the inclusion of data satisfying the condition of 'appearing together with an incorrect answer at $step_{t<i}$' affects the final training outcomes. 

Our results in Tab. \ref{tab:V1V2} indicate that while V2-data yields greater length compression and efficiency in most cases, it comes at the cost of a significant decrease in accuracy. We hypothesize that this is because cases that meet the conditions are more likely to contain thoughtful information, while data that does not meet the aforementioned conditions, in some cases due to the questions being too simple, allows the model to arrive at the correct answer without extensive thinking. In this situation, constructing preference data has led the model to learn a certain degree of lazy thinking habits. We argue that in length compression training, simultaneously improving efficiency and accuracy is of greater value. Consequently, we report the results obtained after training with V1-data. Nevertheless, we acknowledge that V2-data is capable of achieving length compression and efficiency that surpass the baseline.

\subsection{Case Study}
\label{sec:case_study}
\begin{tcolorbox}[
  colback=gray!6,
  colframe=gray!50,
  boxrule=0.4pt,
  arc=2pt,
  left=6pt,
  right=6pt,
  top=6pt,
  bottom=6pt
]
\label{fig:case study}
\textbf{System:} Answer the question.\par
\textbf{User:} Question: A math teacher had \$100 to buy three different types of calculators. A basic calculator costs \$8. A scientific calculator costs twice the price as the basic while a graphing calculator costs thrice the price as the scientific. How much change did she receive after buying those three different types of calculators?\par
\textbf{Probe at $Step_0$}: The answer is  0.\\
\textbf{Probe at $Step_1$}: First, find the cost of each calculator. The answer is  100.\\
\textbf{Probe at $Step_2$}: First, find the cost of each calculator.

- Basic calculator: \$8
- Scientific calculator: 2 * \$8 = \$16
- Graphing calculator: 3 * \$16 = \$48

The answer is  100 - (8 + 16 + 48) = 100 - 72 = \$28.\\
\textbf{Probe at $Step_3$}: First, find the cost of each calculator.

- Basic calculator: \$8
- Scientific calculator: 2 * \$8 = \$16
- Graphing calculator: 3 * \$16 = \$48

Next, add the costs of the three calculators: \$8 + \$16 + \$48 = \$72.
The answer is  72.\\
\textbf{Probe at $Step_4$}: First, find the cost of each calculator.

- Basic calculator: \$8
- Scientific calculator: 2 * \$8 = \$16
- Graphing calculator: 3 * \$16 = \$48

Next, add the costs of the three calculators: \$8 + \$16 + \$48 = \$72.
The teacher had \$100. Subtract the cost of the calculators from the initial amount: \$100 - \$72 = \$28
The answer is  28.
\end{tcolorbox}
\begin{table}[htbp]
\centering
\caption{V1 represents the length preference data at $step_i$ constructed under the condition that "there is an incorrect answer at $step_{t<i}$", while V2 represents data that does not need to satisfy this condition. The bold text represents the best result for the corresponding metric in the block.}
\label{tab:V1V2}
\begin{tabular}{l|lll|lll}
\toprule
\multirow{2}{*}{Method} & \multicolumn{3}{c|}{GSM8K} & \multicolumn{3}{c}{Math} \\
                        & Acc    & Len     & Eff    & Acc    & Len    & Eff    \\
\midrule
BON1                    & 81.6   & 154.0   & 0.530  & 46.3   & 418.5  & 0.111  \\
SKIP1-V1                & \textbf{82.9}   & 115.2   & 0.720  & \textbf{47.8 }  & 398.2  &\textbf{ 0.120}  \\
SKIP1-V2                & 82.1   & \textbf{101.7}   & \textbf{0.807}  & 43.7   & \textbf{380.6}  & 0.115  \\
\midrule
BON8                    & 83.4   & 127.1   & 0.656  & 45.1   & 374.3  & 0.120  \\
SKIP8-V1               & \textbf{84.4 }  & 119.0   & 0.709  & \textbf{46.7}   & 367.9  & 0.127  \\
SKIP8-V2                 & 82.4   & \textbf{97.6}    & \textbf{0.844}  & 45.4   & \textbf{338.0}  & \textbf{0.134 } \\
\bottomrule
\end{tabular}
\end{table}
\begin{table}[htbp]
\centering
\caption{Comparison of inference and training data volumes between BON-8 and SKIP-8 on the Llama3.1-8B-instruct model.}
\label{tab:dataset}
\begin{tabular}{c|cc|cc}
\toprule
\multirow{2}{*}{dataset} & \multicolumn{2}{c|}{BON-8}       & \multicolumn{2}{c}{SKIP-8}                                            \\
                         & \multicolumn{1}{l}{GEN} & TRAIN & \multicolumn{1}{l}{GEN-SFT/DPO} & \multicolumn{1}{l}{TRAIN-SFT/DPO} \\
                         \midrule
GSM8K                    & 7473*8                  & 7473  & 3737*8/3736                     & 3737/3169                         \\
Math                     & 3750*8                  & 3750  & 3750*8/3750                     & 3750/2083                         \\
MedQA                    & 3000*8                  & 3000  & 3000*8/3000                     & 3000/1136              \\
\bottomrule
\end{tabular}
\end{table}
We present a detailed case study here. By inserting probes at the conclusion of various sentences within a mathematical reasoning chain, we observe that the probes yield correct results at both $step_2$ and $step_4$. Notably, the probe at $step_2$ produces a precise yet more concise logical derivation, which validates the efficacy of our probing setup. In contrast, $step_3$ results in an erroneous output because the preceding sentence disrupts the model’s ability to synthesize prior context. When the incorrect response at $step_3$ is paired with the correct one at $step_2$ to construct length preference data, the model learns to truncate reasoning and provide concise answers when the logic is already sufficient. Conversely, when paired with the correct output from $step_4$ as knowledge preference data, the model is encouraged to refine its reasoning until a valid conclusion is reached.

\subsection{Training Dataset Statics}
\label{dataset}
We present in Tab.~\ref{tab:dataset} a comparison of the inference and training data volumes between BON-8 and SKIP-8 across different datasets, using the Llama3.1-8B-instruct model.
Since the preference dataset is constructed by inserting probes during linear generation, each question can be approximately treated as one generation.
When constructing preference pairs, a large amount of data is filtered out because it does not satisfy our predefined rules.
It can be observed that SKIP-8 has lower overall sampling cost and training cost compared to BON-8.

\end{document}